\documentclass[10pt,twocolumn]{article}
\usepackage[margin=0.75in]{geometry}
\usepackage{graphicx}
\usepackage{booktabs}
\usepackage{amsmath}
\usepackage{caption}
\usepackage[hidelinks]{hyperref}
\usepackage{xcolor}
\usepackage{times}
\usepackage{authblk}
\usepackage{balance}

\title{\vspace{-2em}\textbf{Time Capsule of Testable Human Knowledge:\\
41 Years of \textit{Jeopardy!}\ in a Single Free Local Model}}

\author[1]{David Noever}
\author[1]{Forrest McKee}
\affil[1]{PeopleTec, Inc., 4901-D Corporate Drive, Huntsville, AL 35805, USA}
\affil[ ]{\texttt{\{david.noever, forrest.mckee\}@peopletec.com}}
\date{}

\begin{document}
\maketitle

\begin{abstract}
\noindent
In 2011, IBM's Watson was something like a sealed capsule of its era's queryable knowledge. Its DeepQA system defeated the strongest human \textit{Jeopardy!}\ champions, but the knowledge that let it do so lived in a curated billion-document corpus running on a cluster of POWER7 servers, frozen at build time and impossible to move or copy. We show that the same kind of artifact, a snapshot of what a culture can answer, is now portable and essentially free. We evaluate a single 9\,GB open-weight model (Qwen2.5-14B, 4-bit) against the complete open \textit{Jeopardy!}\ clue dataset, 529{,}939 clues across all 41 broadcast seasons from 1984 to 2025. To our knowledge this is the first time a model has been run over the full corpus. The 41 years mark only how long the questions were collected. What they test is far older and broader: the accumulated body of human general knowledge a culture considers worth knowing, from ancient history and dead languages to science, literature, and geography, with a verified answer for every item. The model answers 67.0\% of all clues under a strict forced-response protocol with exact and fuzzy matching, and exceeds 85\% on factoid categories. We treat training-data exposure as something both systems share rather than a flaw unique to language models. Watson's case is in fact the more extreme one. Its corpus was assembled to contain Jeopardy answers and it was tuned on past clues, and it could not answer anything outside that curated distribution. The decisive test is whether a model can answer clues that did not exist when it was built. On clues aired after its training cutoff, the local model holds 65\% and Claude Opus~4.8 holds 95\%, while Watson by construction scores zero. The capability survives the move from a server room to a file you could seal in a time capsule, and unlike Watson it is not frozen to its own moment.
\end{abstract}

\section{Introduction}

The 2011 \textit{Jeopardy!}\ contest between IBM's Watson and champions Ken Jennings and Brad Rutter is rightly remembered as a milestone in artificial intelligence \cite{ferrucci2010,ferrucci2012}. Watson's DeepQA architecture generated candidate answers from a curated corpus of roughly a billion documents, weighed evidence with more than a hundred scoring components, and produced calibrated confidence estimates that governed when it chose to respond \cite{ferrucci2010,fan2012}. It ran on ninety POWER7 servers. It was a remarkable feat of systems engineering, built for one task, and it deserves to be judged by the standards of its era rather than dismissed from a faster-moving present.

The decade since produced a different kind of technology. Transformer-based language models pick up broad world knowledge during general pre-training, with no task-specific retrieval pipeline, and answer in a single forward pass. By 2023, O'Leary \cite{oleary2023} showed that GPT-3.5 and BARD matched or beat Watson on the original contest questions. Those were large hosted systems, though. Our question is one of cost: how much of Watson's question-answering ability now fits in a model small enough to run on a single consumer GPU? A natural objection is that the model may just be repeating clues it saw in pre-training. We take that seriously, but it cuts the other way. Prior exposure to the source material is not a flaw peculiar to language models. It is the condition under which Watson itself operated, in a stricter form, and the real question is which system can answer beyond the material it absorbed.

This paper makes four contributions. We report the first benchmark, as far as we know, of a language model against the complete 41-season, 529{,}939-clue \textit{Jeopardy!}\ corpus. We analyze the full run by difficulty, question type, and individual broadcast year, with very large per-cell samples, and we show across a range of open transformer models from 4.7 to 19\,GB that the result is a property of the model class rather than one checkpoint. We treat training-data exposure symmetrically, arguing it is something both Watson and modern models share and that Watson exhibits more acutely, and we add a post-cutoff held-out test that measures out-of-distribution reach instead of policing contamination. And we reframe the ``Watson challenge'' from a contest of dominance into a measurement of knowledge compression.

\section{Background}

\subsection{The \textit{Jeopardy!}\ task and Watson's design}
\textit{Jeopardy!}\ presents clues as declarative statements within labeled categories, and responses are phrased as questions. Clue dollar values rise with intended difficulty, and the game includes wagering Daily Double and Final Jeopardy clues. The format is linguistically adversarial, dense with wordplay and misdirection \cite{oleary2023}. One defining feature of Watson was selective response. It computed a confidence factor and answered only above a learned threshold, trading coverage for precision \cite{ferrucci2010}. In O'Leary's analysis, Watson answered 75.4\% of contest clues correctly when scored against the full set, but 94.1\% of the high-confidence subset it chose to attempt \cite{oleary2023}. These two numbers measure different things, and we keep them apart.

\subsection{Question-answering benchmarks, then and now}
\textit{Jeopardy!}\ sits in a long line of question-answering benchmarks, and its arc is instructive precisely because it bends toward triviality. Efforts to standardize QA evaluation have wrestled for years with how to score open-ended answers and how to keep a benchmark meaningful as systems improve \cite{usbeck2019,chen2019}. The recurring problem is saturation: a benchmark that distinguishes systems on release can become a solved task within a few model generations, which is the fate we document here for \textit{Jeopardy!}. The field's response has been to build harder targets on purpose. Humanity's Last Exam assembles expert-level questions specifically chosen to resist current frontier models \cite{phan2025}, and parallel efforts extend that idea to specialized domains such as medicine \cite{gallifant2025}. Others probe the format itself, asking whether models can even author questions hard enough to stump themselves \cite{balepur2025}, and the \textit{Jeopardy!}\ format has been extended to other languages, such as a Russian ``Own Game'' question-answering corpus \cite{mikhalkova2022}. Our contribution is the complementary bookend. Rather than constructing a benchmark at the frontier, we measure exhaustively how far a once-formidable benchmark has fallen, using the complete historical corpus rather than a sample, and we use the temporal structure of that corpus to separate genuine capability from memorized recall.
It is worth recalling, without diminishing either achievement, that Deep Blue's 1997 chess victory was eventually understood as fast, deep search over a hand-tuned evaluation function rather than human-like strategic insight. Watson's DeepQA was likewise a sophisticated retrieval-and-scoring engine, with its capability living in explicit, task-specific machinery. Modern language models invert that arrangement and compress the relevant knowledge into their parameters. This study measures how compact that compression has become. The point is efficiency, not a verdict on the earlier systems.

\section{Method}

\subsection{Dataset}
We use the open \textit{Jeopardy!}\ clue dataset \cite{jwolle}, comprising 529{,}939 clues across all 41 broadcast seasons (September 1984 through 2025). Each record contains the round, clue value, Daily-Double value, category, the clue text, the canonical response, and the air date. Pre-2001 dollar values, which were on a half-scale, are normalized when assigning difficulty tiers.

\subsection{Difficulty and type tagging}
Dollar values provide a built-in difficulty grading. We map normalized values to \textsc{easy}, \textsc{medium}, and \textsc{hard} tiers, with \textsc{daily-double} and \textsc{final} kept as separate categories. Separately, we assign each clue a question \emph{type} (person, place, science, definition, date/year, quote/fill-in, wordplay, general) using a lightweight regular-expression taxonomy. The full-coverage run scores every clue, so type and difficulty cells contain tens of thousands of examples each, while the stratified multi-model run of Section~\ref{sec:multimodel} balances cells across these slices for fair per-model comparison.

\subsection{Model and infrastructure}
We evaluate Qwen2.5-14B-Instruct, 4-bit quantized ($\sim$9\,GB), served locally through Ollama with 32-way request concurrency. The complete 529{,}939-clue sweep ran in about 4.6 hours on a single workstation-class GPU, at a sustained median latency of 0.83\,s per clue. That is well within Watson's three-second response window, on hardware orders of magnitude cheaper than a server cluster. We used greedy decoding (temperature 0) for reproducibility and checkpointed results continuously so the run could resume after interruption.

The accessibility of this setup is part of the point. A moderate cloud GPU of the kind offered on free or low-cost tiers, paired with one of the many open small language models now available in this size class, passes through 41 years of the show's clues in a few hours. There is no human equivalent to this throughput. The closest comparison is not a single expert but a crowd-sourced test pooling millions of people, which is roughly what the show itself represents across four decades of contestants. We make the per-person comparison concrete in Section~\ref{sec:human}.

\subsection{Scoring}
A response is correct if, after normalization (lower-casing, article and interrogative-prefix stripping, punctuation removal), it matches the canonical answer under exact or fuzzy token-set criteria at a threshold of 88. We report this lexical accuracy as our primary metric because it is fully reproducible. It is also a conservative lower bound, since it can reject answers that are semantically correct but lexically different. An optional LLM-judge rescue pass recovers a small additional fraction, around 3 to 5\% in prior stratified work, but it is not needed for the headline figures and we omit it here to keep the metric judge-independent.

\section{Results}

\subsection{Full-corpus accuracy}
Across all 529{,}939 clues, Qwen2.5-14B reaches 67.0\% lexical accuracy under forced response. Lexical matching understates true correctness, and Watson's comparable all-clue figure was 75.4\% with a far larger and costlier system. A single 9\,GB model reproduces most of Watson-class performance at a small fraction of the resources.

\subsection{A class of models, not one model}
\label{sec:multimodel}
\begin{figure}[t]
\centering
\includegraphics[width=\linewidth]{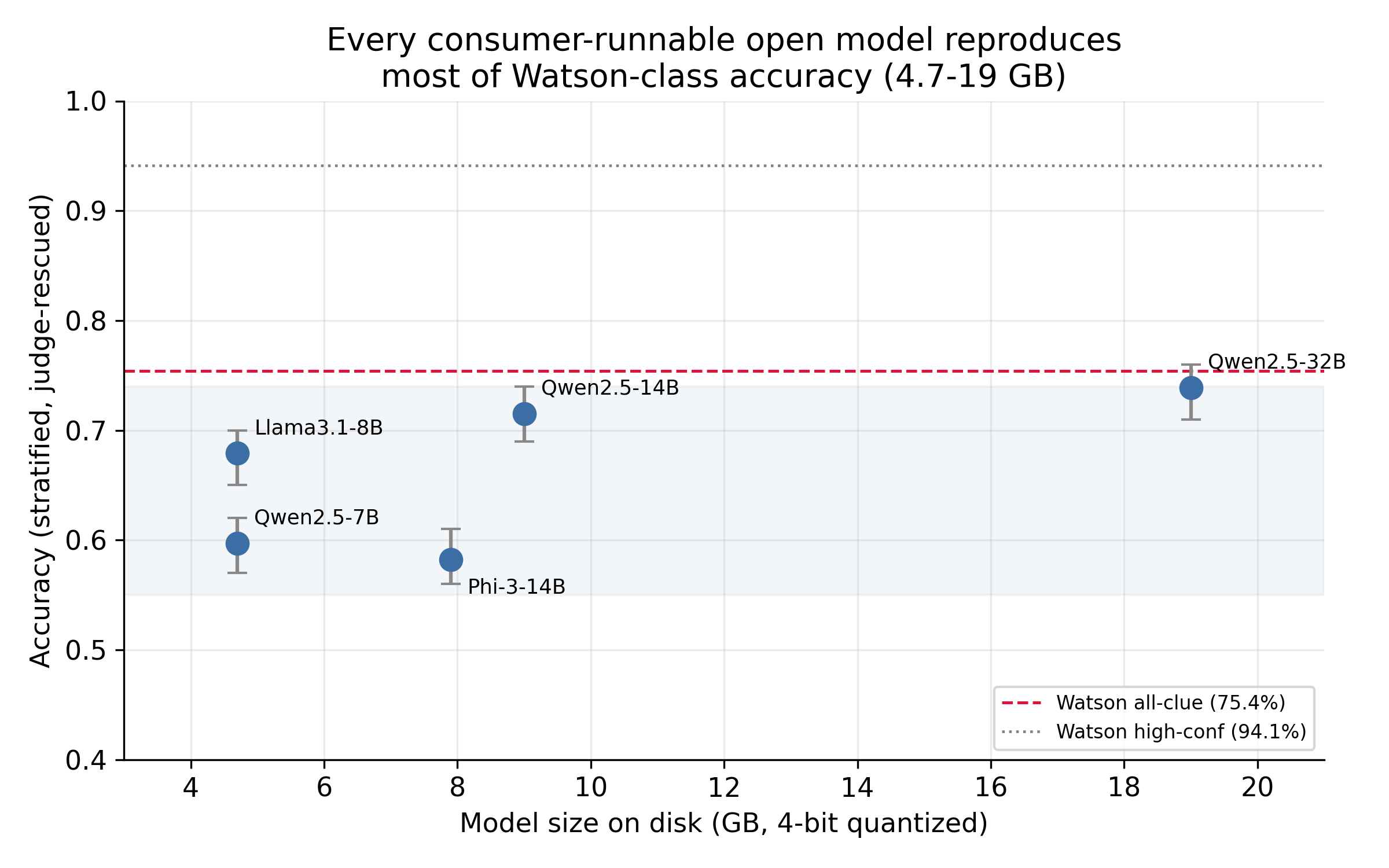}
\caption{Accuracy against model size for five open transformer models (stratified 1{,}384-clue sample, judge-rescued), with Watson reference lines. The takeaway is not that size wins, since within this band the relationship is loose and Phi-3-14B trails much smaller models, but that every model in the consumer-runnable range reproduces most of Watson-class accuracy.}
\label{fig:multimodel}
\end{figure}

The full-corpus sweep uses a single model for tractability, but the finding is not specific to it. In a companion stratified run we scored five open transformer models spanning roughly 4.7 to 19\,GB on disk (4-bit) against a balanced 1{,}384-clue sample drawn across difficulty and question type, with judge-rescued scoring. Every model in the band answers a clear majority of clues: Qwen2.5-7B at 59.7\%, Phi-3-14B at 58.2\%, Llama3.1-8B at 67.9\%, Qwen2.5-14B at 71.5\%, and Qwen2.5-32B at 73.9\%, the last within a point of Watson's 75.4\% all-clue figure (Fig.~\ref{fig:multimodel}).

We are careful about what this does and does not show. It is not a scaling-law claim. The point is not that larger wins, and within this narrow band size predicts accuracy only loosely, with Phi-3-14B sitting below much smaller models because its training mixture favors reasoning and code over broad trivia. The point is the opposite and more striking one: across architectures, vendors, and a four-fold spread in size, every model in the consumer-runnable range reproduces most of Watson-class question answering. The remarkable fact is not the few points that separate these models from each other but that a 9\,GB file, a roughly 8{,}000-fold compression of its training text (Section~\ref{sec:capsule}), retains this much of a civilization's general knowledge at all. Competence here is a property of the current generation of open models as a class, not a quirk of one checkpoint, which is what lets us treat the single full-corpus model as representative. Per-difficulty, per-type, and per-decade breakdowns for all five models are given in Appendix~\ref{app:multimodel}.

\subsection{Difficulty}
\begin{figure}[t]
\centering
\includegraphics[width=\linewidth]{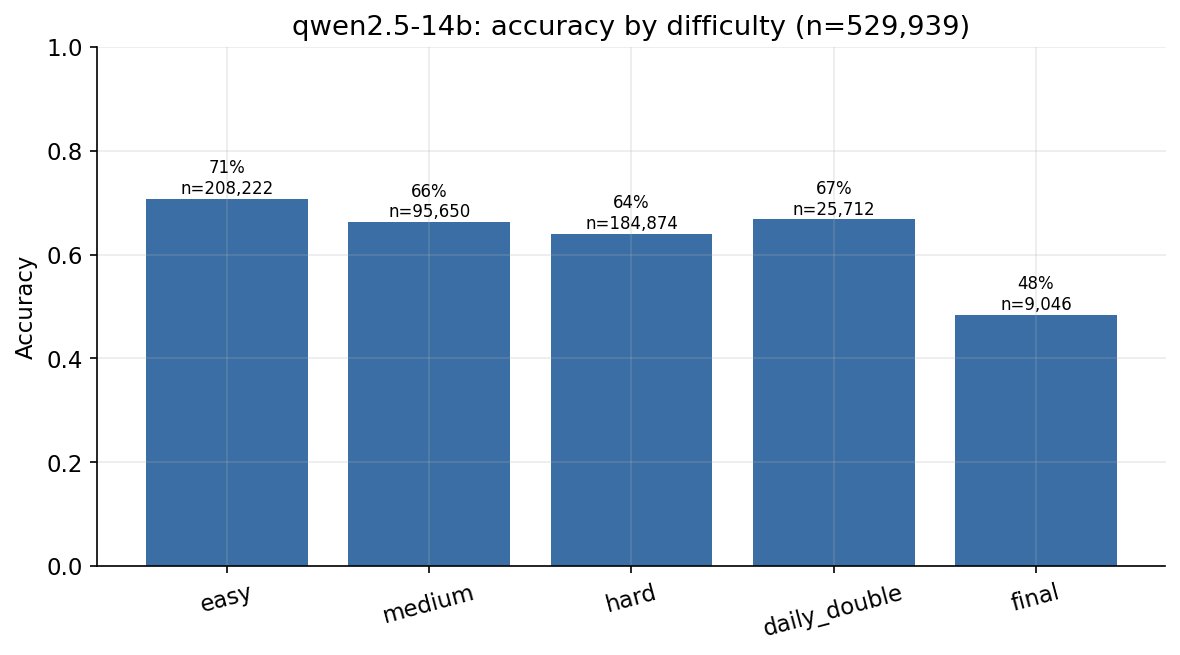}
\caption{Accuracy by difficulty tier across the full 529{,}939-clue corpus. Bars annotated with per-tier sample sizes. Only \textsc{final} clues fall markedly.}
\label{fig:difficulty}
\end{figure}

Figure~\ref{fig:difficulty} shows accuracy by difficulty tier with the full-corpus sample sizes. Performance is nearly flat across \textsc{easy} (71\%), \textsc{medium} (66\%), \textsc{hard} (64\%), and \textsc{daily-double} (67\%), and falls sharply only on \textsc{final} clues (48\%), which are deliberately the most lateral. Dollar value is a proxy for human difficulty, and the fact that it barely predicts model accuracy echoes O'Leary's finding for ChatGPT and BARD \cite{oleary2023}.

\subsection{Question type}
\begin{figure}[t]
\centering
\includegraphics[width=\linewidth]{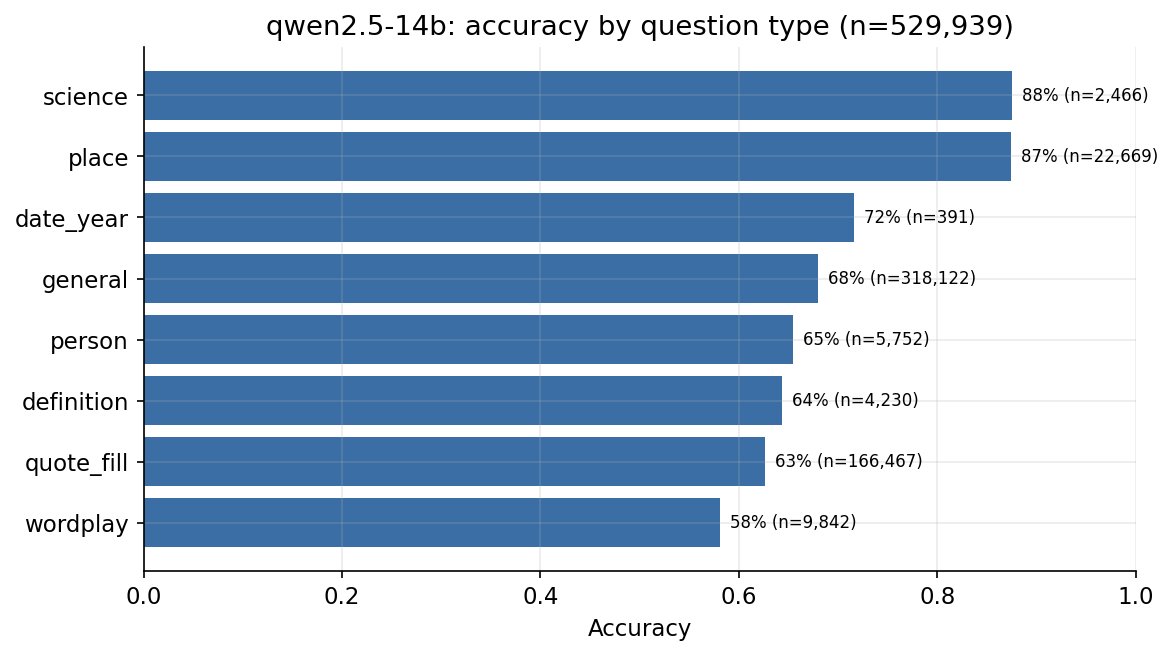}
\caption{Accuracy by question type across the full corpus, with per-type $n$. Factoid types (science, place) lead, and wordplay trails.}
\label{fig:qtype}
\end{figure}

Figure~\ref{fig:qtype} reports accuracy by type. The model does best on \textsc{science} (88\%) and \textsc{place} (87\%), both factoid categories with crisp canonical answers, and worst on \textsc{wordplay} (58\%), which rewards the surface-form, lateral reasoning that compact models find hardest.

\subsection{The year curve and shared exposure}
\begin{figure}[t]
\centering
\includegraphics[width=\linewidth]{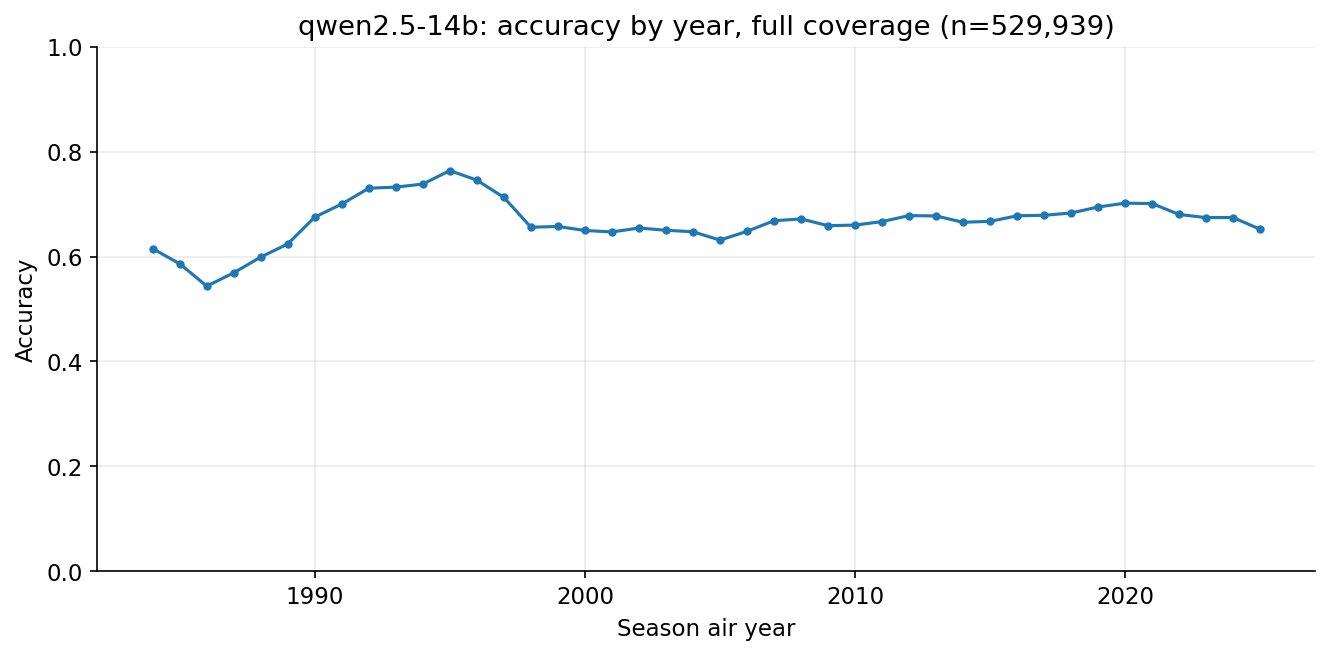}
\caption{Accuracy by individual broadcast year (full coverage). Flatness reflects knowledge of historical material absorbed during pre-training, the same kind of prior exposure Watson's curated corpus depended on. Whether the model generalizes beyond that material is tested separately (Sec.~\ref{sec:ood}).}
\label{fig:year}
\end{figure}

Figure~\ref{fig:year} plots accuracy for each individual broadcast year. Accuracy is essentially flat from the late 1990s through 2025, sitting near 65 to 70\% with a mild early-season bump. One reading is that the model has memorized clues that appear in its pre-training data through the open dataset and its many copies. That is probably true to some degree, but it is not a weakness that sets the model apart. Watson rested on the same foundation in a more deliberate form. Its corpus was hand-curated to contain the answers to Jeopardy clues, and its models were tuned on thousands of past clues \cite{ferrucci2010,fan2012}. Prior exposure to the historical material is the shared substrate of both systems, not a contaminant of one. What the flat curve cannot settle on its own is whether the model can answer outside that absorbed distribution. Watson's architecture answered that in the negative by construction, and we test it directly in the next section.

\section{Out-of-Distribution Reach}
\label{sec:ood}

The decisive difference between Watson and a modern language model is not how much each absorbed from the historical record. Both absorbed a great deal. It is whether either can answer questions outside what it absorbed. Watson could not. Its corpus and scoring models were fixed at build time, which bounded its competence by construction. We test whether the language model can, using held-out clues it could not have seen during training.

\textbf{Condition A (in-distribution).} The full 530k sweep above, at 67.0\%. This measures performance on material the model plausibly saw in pre-training, the regime in which Watson also operated.

\textbf{Condition B (post-cutoff).} Clues from seasons aired after the model's knowledge cutoff, 2025 onward for Qwen2.5. These clues did not exist when the model was trained, so accuracy here cannot come from having seen the clue. The gap $A-B$ is not a contamination penalty. It measures out-of-distribution reach, the extent to which the model answers clues it could not have memorized. A small gap is the interesting result, because it shows the capability is transferable knowledge rather than clue-specific recall. This is the regime Watson could never enter.

\textbf{Condition C (frontier ceiling).} A current frontier model (Claude Opus~4.8) on a small random sample, run on both in-distribution and post-cutoff clues, establishes how high the ceiling sits and whether the frontier's advantage holds up on held-out material.

\subsection{Results across conditions}
\begin{figure}[t]
\centering
\includegraphics[width=\linewidth]{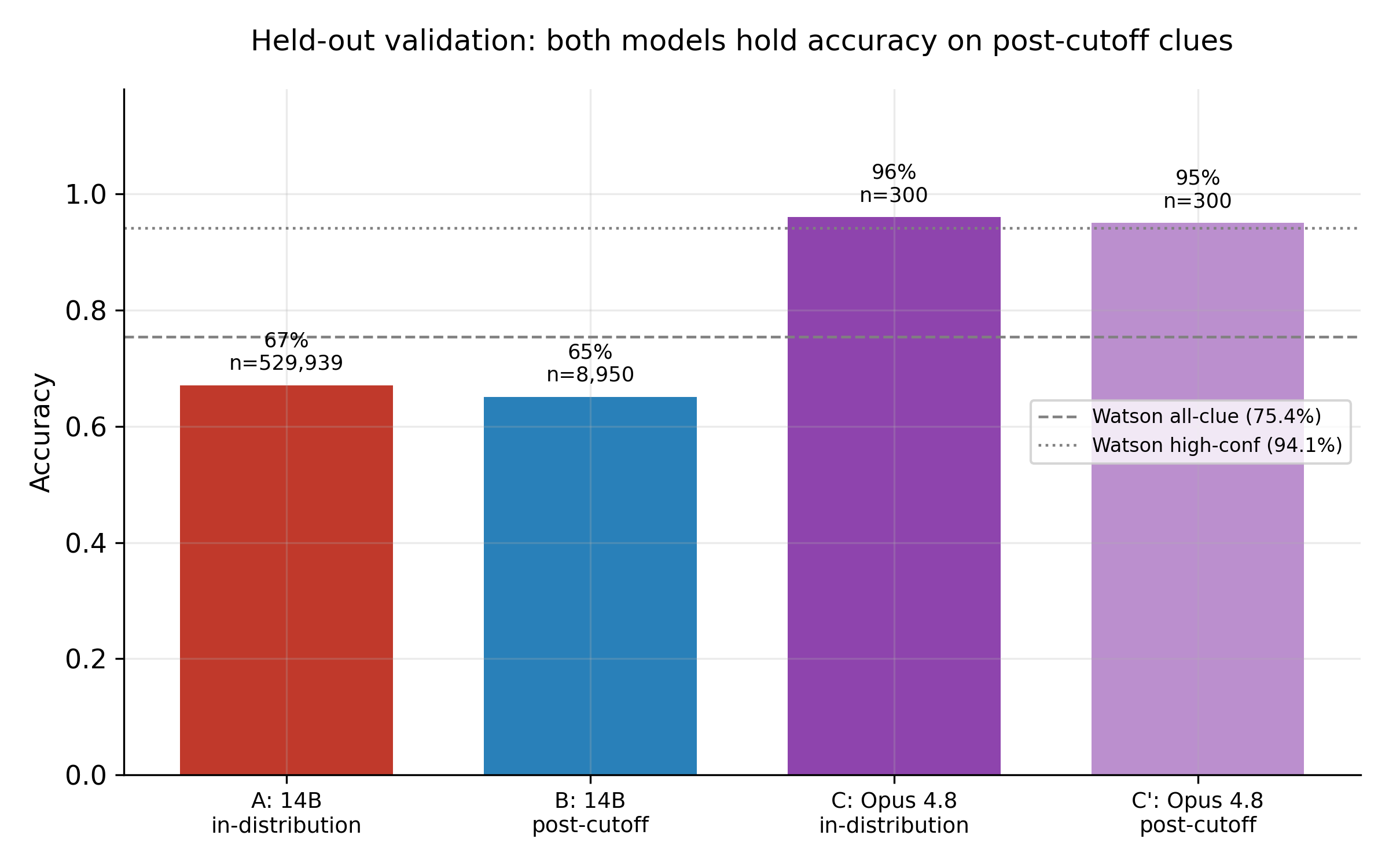}
\caption{Accuracy across conditions, against Watson reference lines. For both the local 14B and Claude Opus~4.8, the post-cutoff bar barely trails in-distribution, which indicates the score reflects transferable knowledge rather than clue-specific memorization.}
\label{fig:leakage}
\end{figure}

Table~\ref{tab:leakage} and Figure~\ref{fig:leakage} give accuracy across the three conditions. The result that matters is how little the held-out test costs either model. The local 14B answers 67.0\% of in-distribution clues and 65\% of post-cutoff clues, a gap of two points. Claude Opus~4.8 answers 96\% in-distribution and 95\% post-cutoff, a gap of one point. Neither model leans on having seen the specific clue. If performance were mostly memorized recall, accuracy would fall sharply on clues that aired after training ended, and it does not. The capability is transferable knowledge, which is the regime Watson could not enter at all, since its corpus was fixed at build time.

\begin{figure}[t]
\centering
\includegraphics[width=\linewidth]{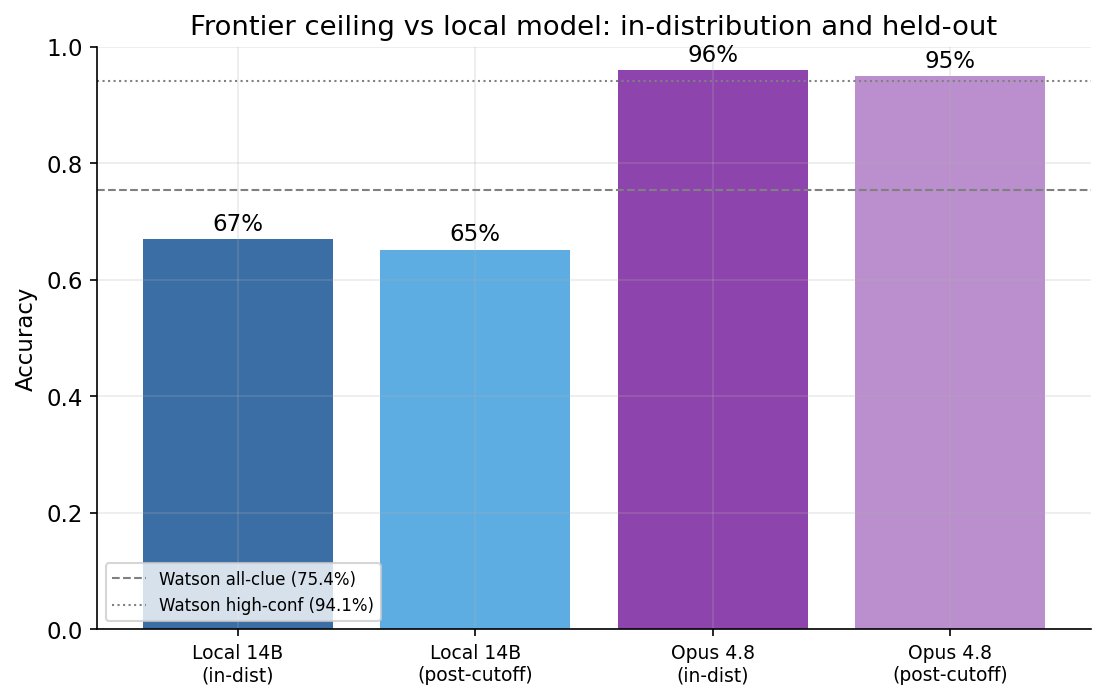}
\caption{Local 14B versus Claude Opus~4.8, each on in-distribution and post-cutoff clues. Both models hold their accuracy on held-out material (67$\to$65\% and 96$\to$95\%), the signature of transferable knowledge rather than memorized clues.}
\label{fig:frontier}
\end{figure}

Two further points stand out in Figure~\ref{fig:frontier}. The frontier model sits near the ceiling on both conditions, just under Watson's 94.1\% high-confidence precision but on every clue rather than a self-selected subset. And the small local model, at 65\% on genuinely unseen clues, already clears the range where pre-Watson QA systems operated and lands close to Watson's own 75.4\% all-clue figure, on hardware that fits in a backpack.

\begin{table}[t]\centering\small
\caption{Held-out validation. A=in-distribution, B=post-cutoff (both Qwen2.5-14B). C and C$'$ are Claude Opus~4.8 on the same two splits.}\label{tab:leakage}
\begin{tabular}{llcc}
\toprule
Cond. & Description & $n$ & Accuracy \\
\midrule
A    & 14B, in-distribution      & 529{,}939 & 0.670 \\
B    & 14B, post-cutoff          & 8{,}950   & 0.650 \\
C    & Opus 4.8, in-distribution & 300       & 0.960 \\
C$'$ & Opus 4.8, post-cutoff     & 300       & 0.950 \\
\bottomrule
\end{tabular}
\end{table}

\section{Why the Benchmark Aged}
\label{sec:aged}

It is tempting to read these results as ``models got big enough to memorize trivia,'' but that misreads what made \textit{Jeopardy!}\ hard in 2011 and why it is easier now. The difficulty Watson confronted was only partly about knowing facts. A large share of its engineering went toward language understanding, and that is the capability the field has solved most thoroughly in the fifteen years since.

\subsection{What was hard in 2011}
A \textit{Jeopardy!}\ clue is not a plainly phrased question. It is a declarative statement, often syntactically convoluted, that buries the information need inside misdirection and wordplay and leaves the required response form (who, what, where, when) implicit. DeepQA handled this with explicit machinery: components for question-type classification, for parsing the clue's syntax into a literal information need, and for the puns and lexical games that \textit{Jeopardy!}\ writers favor \cite{ferrucci2010,fan2012}. In 2011, working out what a clue was even asking was a substantial problem that had to be engineered separately. The linguistic form of the clue was itself an adversary.

\subsection{What instruction-tuned models get for free}
An instruction-tuned language model in 2026 dissolves most of that layer. Mapping an oblique, convoluted phrasing onto the underlying intent is the central thing instruction-tuning produces. The model takes in the clue and category and behaves as though it just parses what is being asked, with no question-type classifier or syntactic pipeline behind it. Our data show this directly. Clue dollar value, which is the show's own difficulty grading and largely a function of phrasing complexity, barely predicts accuracy. Performance is nearly flat across the \textsc{easy} (71\%), \textsc{medium} (66\%), \textsc{hard} (64\%), and \textsc{daily-double} (67\%) tiers (Fig.~\ref{fig:difficulty}). The convolution built to challenge human contestants, and which Watson had to parse explicitly, scarcely registers.

\subsection{What remains genuinely hard}
The benchmark did not become uniformly trivial. Its difficulty migrated. Two categories resist the model because they require manipulating the surface form of the words rather than retrieving knowledge about the world. \textsc{Wordplay} clues, which turn on anagrams, hidden words, and rhyme- and letter-based puzzles, are the lowest-scoring type at 58\% (Fig.~\ref{fig:qtype}), well below factoid categories like science (88\%) and place (87\%). \textsc{Final Jeopardy} clues, the most lateral by design, fall to 48\%. These residuals tell us something. Instruction-tuned models have largely solved the parsing of indirect phrasing, the problem of understanding what a convoluted clue asks, while only partly solving the resolution of genuine wordplay, where the answer depends on operating over the letters and sounds of the clue itself. Watson struggled with both halves. Modern models have effectively closed the first and only dented the second.

\subsection{The implication}
The right conclusion is narrower and more interesting than ``trivia is solved.'' The specific difficulty that made \textit{Jeopardy!}\ an NLP grand challenge, linguistic indirection layered over open-domain recall, split into two parts that have aged very differently. The comprehension half, which was much of Watson's explicit engineering, is now a default property of general models and is effectively free. The symbolic-manipulation half is still a real frontier, even if a narrow one. A benchmark ages not when models simply get bigger, but when the particular capability it was built to stress becomes commodity. \textit{Jeopardy!}\ aged because its central demand, robustly understanding what an awkwardly worded question wants, is exactly what instruction-tuning delivers. We mean this as a claim about observable task behavior, not about machine cognition. Whatever the underlying mechanism, models now act as though they parse instruction and indirection robustly, and that is the capability Watson had to build by hand.

\section{Discussion}

\subsection{Compression, not conquest}
The right framing is knowledge compression. Watson encoded its capability in an explicit corpus and pipeline. A 9\,GB model encodes a comparable breadth of trivia knowledge in its weights, retrievable in under a second on commodity hardware. The interesting question is the compression ratio and where it breaks down. The multi-model results of Section~\ref{sec:multimodel} show accuracy rising with size across the 4.7 to 19\,GB range, which suggests the capability degrades gracefully as models shrink rather than collapsing at some threshold. Pinning down how small a model can still rival Watson-class QA, and how that floor moves with architecture and training mixture, is the natural next step.

\subsection{What the capsule actually holds}
\label{sec:capsule}
We can be more precise about the artifact than the Watson comparison alone allows. Watson's knowledge base was described as roughly a billion documents \cite{ferrucci2010}, but its physical footprint as a running system, spread across a server cluster, was never a portable quantity. The model we use is exactly specified. Qwen2.5 was pre-trained on 18 trillion tokens \cite{qwen2025}, on the order of 70\,terabytes of raw text, comparable in scale to several times the digitized text holdings of a national library. The trained 14B model, quantized to 4 bits, is a single 9\,GB file. That is a compression of the training text into weights of roughly 8{,}000 to 1.

Two cautions keep this honest. The model is not a lossless archive of those 18 trillion tokens. It cannot reproduce its training corpus, and it answers from a compressed, lossy, generalizing representation rather than a stored copy. And the 9\,GB figure is the model, not a database of Jeopardy answers. What the file holds is the capacity to answer, demonstrated here across 41 years of clues including ones that postdate it, not a lookup table of the clues themselves. With that understood, the compactness is the striking part. The same broad question-answering competence that in 2011 required a room of servers now sits in a file small enough to attach to an email.

\subsection{Media for a knowledge capsule}
The size makes the time-capsule framing literal rather than rhetorical. A 9\,GB model fits, with room to spare, on a fingernail-sized microSD card, on a single Blu-ray disc, or on a cheap USB stick. It occupies a rounding error of an 18\,TB LTO magnetic tape cartridge, a medium rated for decades of archival stability. Stored in cloud cold storage such as Amazon S3 Glacier Deep Archive, it costs on the order of a cent per month to keep indefinitely. None of these media existed as options for Watson, whose knowledge could not leave the cluster it ran on. A compact model is the first form in which a culture's broad, testable question-answering ability becomes a physical object one could etch onto durable media, seal in a vault, or in principle send aboard a spacecraft, in the spirit of the Voyager record (Fig.~\ref{fig:capsule}).

\subsection{A capsule you can question}
\label{sec:oracle}
What makes the artifact more than a stored record is that it can be queried about things it was never shown. A flat archive of clues and answers would preserve the past as a lookup table, useful only for questions someone already thought to write down. A model answers questions that were never in the corpus, which the post-cutoff result demonstrates directly. This suggests a thought experiment. Imagine the model washes ashore on a flash drive, the only surviving fragment of a technical civilization, and a finder with a small solar panel and a salvaged screen can ask it anything. The value is not in replaying old game-show clues. It is that the finder can pose new, practical questions and get usable answers drawn from the same body of knowledge the clues sampled.

The corpus touches most of what such a finder would need. From its science and nature categories the model can say that boiling water kills the organisms that cause disease, that willow bark contains the compound aspirin is made from, and which common plants and mushrooms are poisonous. From history and geography it can explain finding true north from Polaris, that bronze is an alloy of copper and tin, and that gunpowder combines charcoal, sulfur, and saltpeter. From its coverage of agriculture and medicine it can describe crop rotation, nitrogen-fixing legumes, the waterborne nature of cholera, and the basics of human anatomy. None of these are exotic facts, which is the point. They are the ordinary contents of a culture's general knowledge, the things deemed common enough to ask on a quiz show, and they are exactly the things a survivor would want to recover. A record that can be questioned, rather than only read, turns preserved knowledge back into usable knowledge.

\subsection{The human equivalent}
\label{sec:human}
It helps to put the sweep in human terms. The 529{,}939 clues are roughly 8{,}700 complete games, since a single match has 61 clues across its two rounds and the final. Ken Jennings's celebrated 2004 streak ran 74 games, so the corpus is on the order of a hundred such streaks stacked end to end. A person trying merely to read and answer every clue, at a brisk fifteen seconds each, would need about 2{,}200 hours, or a full year of forty-hour weeks doing nothing else. No human contestant has faced anything close to this breadth, because no one can specialize across every category the show has ever used. The model worked through all of it in about four and a half hours on one GPU, and a frontier model answered a representative sample at 95\% on clues that did not exist during its training. The comparison is not really runtime against runtime. It is a year of sustained human effort, spread across knowledge no single person holds, against an afternoon on hardware that costs less than a used car.

\subsection{Exposure is symmetric, and Watson's was worse}
It is tempting to dismiss a language model's \textit{Jeopardy!}\ performance as memorization of clues it saw in training. Applied evenly, though, that standard indicts Watson far more than its successor. Watson's billion-document corpus was assembled to contain the answers to Jeopardy clues, and its scoring models were trained on thousands of actual past clues \cite{ferrucci2010,fan2012}. Its competence was bounded by that curated distribution by design. The language model's exposure is broader and incidental rather than targeted, and it is not a ceiling. Where Watson could answer only within the material it was built around, the language model answers clues that aired after its training ended (Condition B). Earlier comparisons, including the Watson-vs-BARD-vs-ChatGPT study \cite{oleary2023}, noted possible contamination but could not control for it, since the contest questions predate the models. The full corpus's temporal structure lets us do what they could not. We measure performance on genuinely unseen clues, which isolates the one capability that actually separates the two eras: generalization beyond the absorbed distribution.

\subsection{Limitations}
The forced-response protocol is stricter than Watson's selective buzzing, so our single-model figure understates the precision achievable with a confidence gate. The regular-expression type taxonomy is approximate. Lexical scoring is conservative and judge-independent by design, trading a few points of recall for full reproducibility. The in-distribution sweep reflects knowledge the model absorbed during training, just as Watson's performance reflected its curated corpus. The post-cutoff condition, not the headline 67.0\%, is what isolates generalization beyond that absorbed material, and we weight our conclusions toward it.

\section{Conclusion}
Across all 41 seasons and 529{,}939 clues, a single 9\,GB open-weight model answers two-thirds of historical \textit{Jeopardy!}\ clues under a strict, reproducible protocol, and far more on factoid categories. That approaches the all-clue accuracy of the 2011 Watson system, on hardware that costs a tiny fraction as much. We do not read this as a verdict on Watson, which was a pioneering system and is properly judged by the standards of its time. We read it as a measure of how thoroughly broad question-answering knowledge now compresses into compact, general models. Prior exposure to the historical material is common to both systems, and Watson's was the more deliberate, so the honest comparison turns not on who saw what but on who can answer beyond it. That is a regime the language model enters and Watson's architecture could not.

\begin{figure*}[t]
\centering
\includegraphics[width=0.92\textwidth]{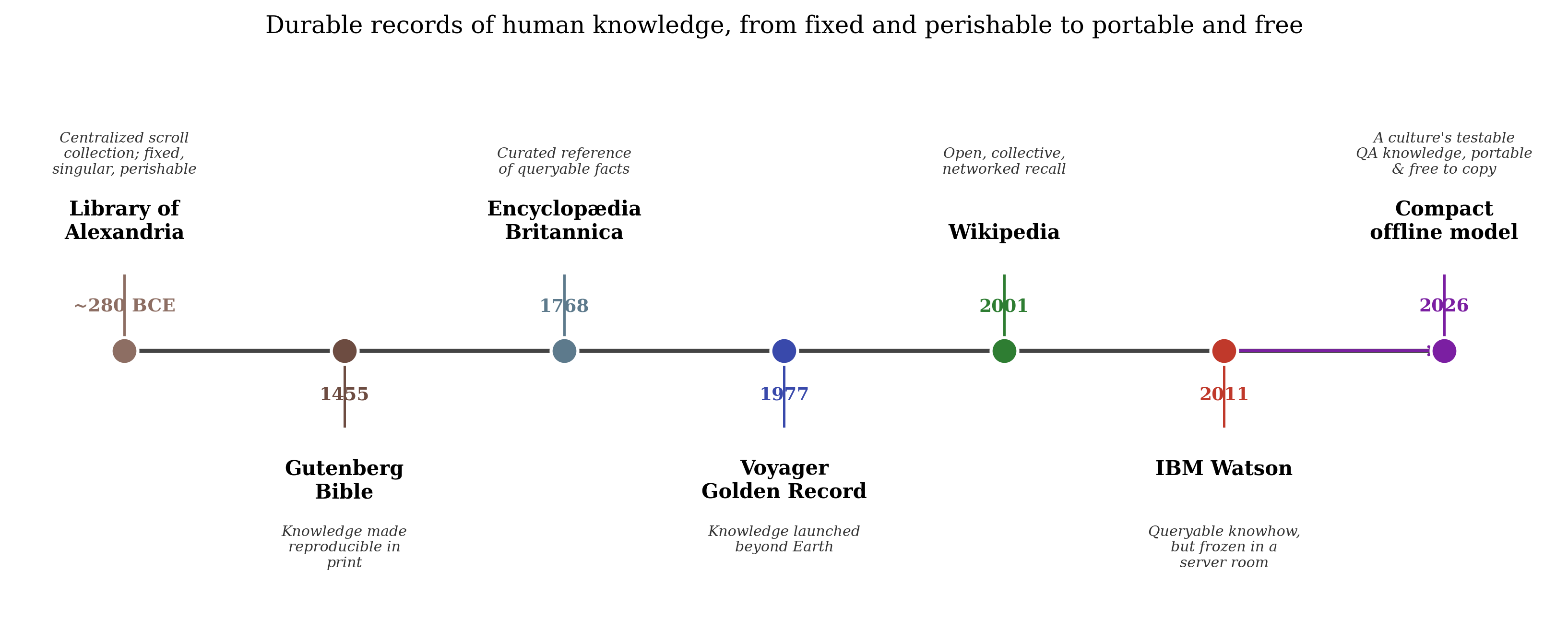}
\caption{Durable records of human knowledge across history. Each marks a shift in how a culture's knowledge is stored and accessed, moving from fixed and perishable collections toward portable, reproducible, and queryable forms. Watson (2011) made a culture's question-answering knowledge queryable but left it frozen in a server room. A compact offline model encodes a comparable record of testable knowledge in a file that can be copied for nothing and carried anywhere.}
\label{fig:capsule}
\end{figure*}

It is worth stepping back from the benchmark to what the artifact is. The corpus is a verifiable record of what a culture holds worth knowing, sampled by 41 years of question writers but reaching across the whole inherited span of human knowledge, every item paired with a known-correct response. In 2011 that record, in queryable form, occupied a server room and was frozen in place. It now fits in a file small enough to bury, mail, or carry, and a model that holds it can answer questions that postdate its own making. Humanity has reached for durable records of its knowledge before, from the Library of Alexandria to the printed Bible to the Voyager golden record (Fig.~\ref{fig:capsule}). A compact offline model that encodes a culture's testable question-answering knowledge, and can be copied for nothing, belongs on that list. The open question is no longer whether a machine can match Watson, but how small and how lasting that capsule can be.

\balance

\onecolumn
\appendix
\section{Multi-model breakdowns}
\label{app:multimodel}

This appendix gives the per-slice detail behind Section~\ref{sec:multimodel}, for all five models on the balanced 1{,}384-clue stratified sample. We summarize the patterns as three observations that hold across the model class rather than reading them as scaling results.

\begin{figure}[h]
\centering
\includegraphics[width=0.62\linewidth]{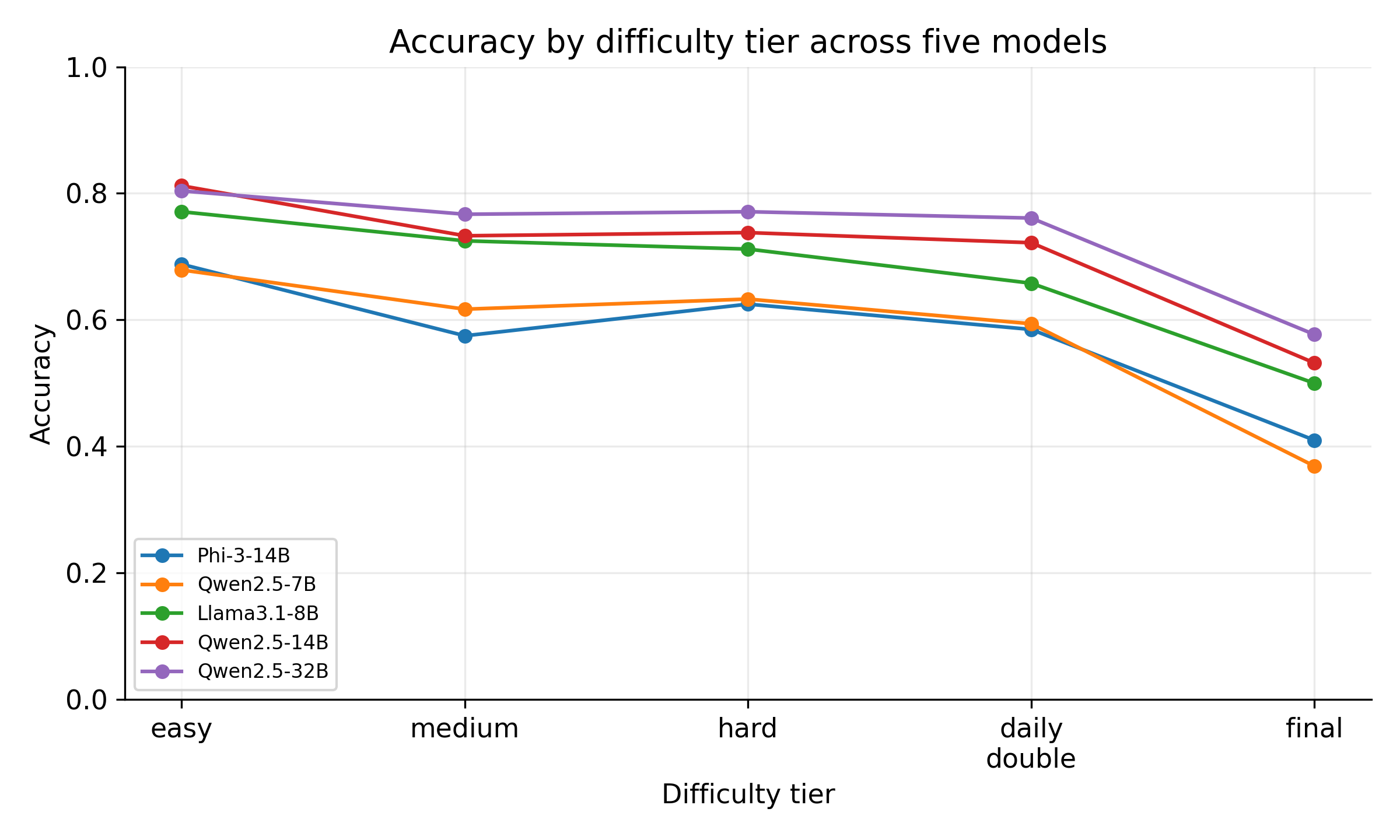}
\caption{Accuracy by difficulty tier for all five models. The shape is shared: a gentle easy-to-hard decline and a sharp drop on \textsc{final} clues.}
\label{fig:app_diff}
\end{figure}

\textbf{Observation 1: difficulty costs all models about the same shape.} Figure~\ref{fig:app_diff} shows accuracy falling gently from \textsc{easy} to \textsc{hard} for every model, then dropping sharply on \textsc{final} clues. The decline tracks the show's own dollar-value grading only weakly until the final tier, where lateral phrasing dominates. The ordering of models is roughly preserved across tiers, so the difficulty effect is a property of the task, not of any one model.

\begin{figure}[h]
\centering
\includegraphics[width=0.62\linewidth]{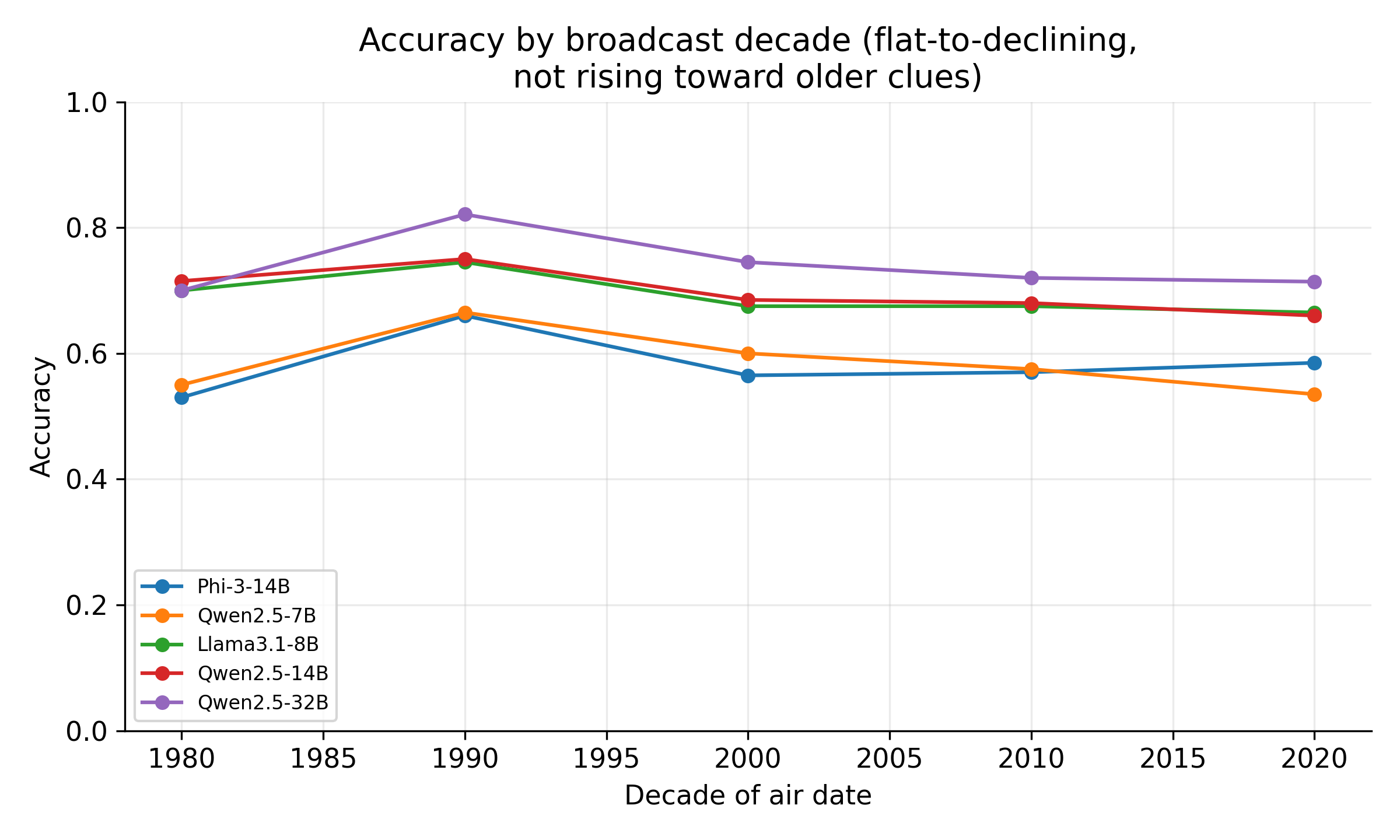}
\caption{Accuracy by broadcast decade for all five models. The flat-to-declining trend across every model argues against memorization of older clues as the dominant mechanism.}
\label{fig:app_dec}
\end{figure}

\textbf{Observation 2: accuracy does not rise toward older clues.} Figure~\ref{fig:app_dec} plots accuracy by broadcast decade. If performance were driven by memorizing clues that have circulated online longest, older decades would score highest. Instead every model is flat or mildly declining from the 1990s to the 2020s. This is the same evidence as the full-corpus year curve, reproduced across the model class, and it points to durable knowledge rather than verbatim recall. Whatever contamination exists applies equally to Watson, which was trained on historical \textit{Jeopardy!}\ material as well.

\textbf{Observation 3: the hard and easy categories are shared.} Table~\ref{tab:app_type} gives accuracy by question type. Factoid categories with crisp answers (science, place) are easiest for every model, and \textsc{wordplay} is hardest for every model. The residual difficulty is surface-form symbol manipulation, consistent with the argument in Section~\ref{sec:aged}. Again the pattern is shared across the class rather than specific to one checkpoint.

\begin{table}[h]\centering\small
\caption{Accuracy by question type across all five models (stratified sample). Factoid types lead and wordplay trails for every model.}
\label{tab:app_type}
\begin{tabular}{lccccc}
\toprule
Type & Phi-3-14B & Qwen2.5-7B & Llama3.1-8B & Qwen2.5-14B & Qwen2.5-32B \\
\midrule
science    & 0.75 & 0.78 & 0.82 & 0.87 & 0.93 \\
place      & 0.78 & 0.82 & 0.87 & 0.86 & 0.88 \\
date/year  & 0.60 & 0.60 & 0.71 & 0.77 & 0.83 \\
general    & 0.54 & 0.58 & 0.67 & 0.74 & 0.66 \\
person     & 0.49 & 0.52 & 0.67 & 0.67 & 0.72 \\
quote/fill & 0.51 & 0.59 & 0.64 & 0.68 & 0.72 \\
definition & 0.55 & 0.50 & 0.55 & 0.62 & 0.64 \\
wordplay   & 0.45 & 0.38 & 0.52 & 0.53 & 0.55 \\
\bottomrule
\end{tabular}
\end{table}

\end{document}